# Transfer Learning With Densenet201 Architecture Model For Potato Leaf Disease Classification

1st Rifqi Alfinnur Charisma
*Faculty Of Informatics*
*Institut Teknologi Telkom Purwokerto*
Purwokerto, Indonesia
19104031@ittelkom-pwt.ac.id

2rd Faisal Dharma Adhinata
*Faculty Of Informatics*
*Institut Teknologi Telkom Purwokerto*
Purwokerto, Indonesia
faisal@ittelkom-pwt.ac.id

*Abstract*— **Potato plants are plants that are beneficial to humans. Like other plants in general, potato plants also have diseases; if this disease is not treated immediately, there will be a significant decrease in food production. Therefore, it is necessary to detect diseases quickly and precisely so that disease control can be carried out effectively and efficiently. Classification of potato leaf disease can be done directly. Still, the symptoms cannot always explain the type of disease that attacks potato leaves because there are many types of diseases with symptoms that look the same. Humans also have deficiencies in determining the results of identification of potato leaf disease, so sometimes the results of identification between individuals can be different. Therefore, the use of Deep Learning for the classification process of potato leaf disease is expected to shorten the time and have a high classification accuracy. This study uses a deep learning method with the DenseNet201 architecture. The choice to use the DenseNet201 algorithm in this study is because the model can identify important features of potato leaves and recognize early signs of emerging diseases. This study aimed to evaluate the effectiveness of the transfer learning method with the DenseNet201 architecture in increasing the classification accuracy of potato leaf disease compared to traditional classification methods. This study uses two types of scenarios, namely, comparing the number of dropouts and comparing the three optimizers. This test produces the best model using dropout 0.1 and Adam optimizer with an accuracy of 99.5% for training, 95.2% for validation, and 96% for the confusion matrix. In this study, using data testing, as many as 40 images were tested into the model that has been built. The test results on this model resulted in a new accuracy for classifying potato leaf disease, namely 92.5%.**

## I. INTRODUCTION

Potatoes are one of the staple foods in Indonesia because they contain lots of carbohydrates [1]. Like other plants in general, potato plants also have diseases; if this disease is not treated immediately, there will be a significant decrease in food production. Therefore it is necessary to detect diseases quickly and precisely so that disease control can be carried out effectively and efficiently [2].

Classification of potato leaf disease can be done directly. Still, the symptoms cannot always explain the type of disease that attacks potato leaves because there are many types of diseases with symptoms that look the same. Humans also have deficiencies in determining the results of identification of potato leaf disease, so sometimes the results of identification between individuals can be different [3][4]. It is caused by reduced concentration and fatigue. Also, it requires a lot of experience in identifying potato leaf diseases; classifying potato leaves in large quantities will undoubtedly take longer [5][6]. Therefore, the use of Deep Learning for the classification process of potato leaf disease is expected to shorten the time and have a high classification accuracy [7][8].

Several studies have classified potato leaf disease, namely [9] using the VGG19 algorithm and [10] using the ResNet-152 and InceptionV3 algorithms. Classification of potato leaf disease has been carried out using different classification algorithms. So this research will use the Transfer Learning classification algorithm with the DenseNet201 architecture. The training uses this method because DenseNet201 is one of the artificial neural network architectures used for image recognition tasks. DenseNet201 uses the concepts of dense block and transition layer to improve image recognition efficiency and reduce overfitting. DenseNet201 can also deal with the vanishing gradient problem often found in complex artificial neural network architectures. In addition, the DenseNet201 model can identify important characteristics of potato leaves and recognize early signs of emerging diseases. This study aims to evaluate the effectiveness of the transfer learning method with the DenseNet201 architecture in increasing potato leaf disease classification accuracy compared to traditional classification methods.

## II. RESEARCH METHODOLOGY

### A. Research Stages

The following is a research flowchart containing the stages of the research to be carried out.

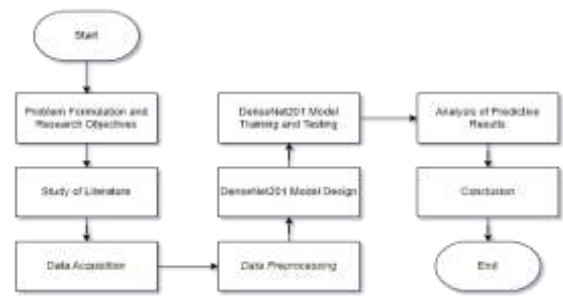

**Fig. 1** Research Flowchart

### B. Problem Formulation and Research Objectives

The formulation of the problem is carried out to find the existing problems so that this research is needed, and the preparation of research objectives is carried out to determine the purpose of this research [11].

### C. Study of Literature

The next stage is library research. At this stage, the author reads and understands deep learning concepts and problems in journals, books, and previous research. Furthermore, the results obtained are used as the basis for writing and research to be carried out [12].

*D. Data Acquisition*

Data collection in this study was carried out quantitatively. Potato leaf image data was obtained through the Kaggle website [13] with 3900 potato leaf images.

*E. Data Preprocessing*

Preprocessing is the initial process of improving an image to remove noise. As stated by [14], preprocessing is a process to remove unnecessary parts of the input image for further processing. The image preprocessing technique in this study uses the ImageDataGenerator library to determine the same shape and size. Later, the image will be rotated, flipped, and resized to a certain scale [15]. In this study, the image is resized to 224x224 pixels.

*F. DenseNet201 Model Design*

In this study, there were several stages to identifying potato leaf disease. This research is a research based on several existing studies. The general steps for identifying potato leaf disease are in figure 2.

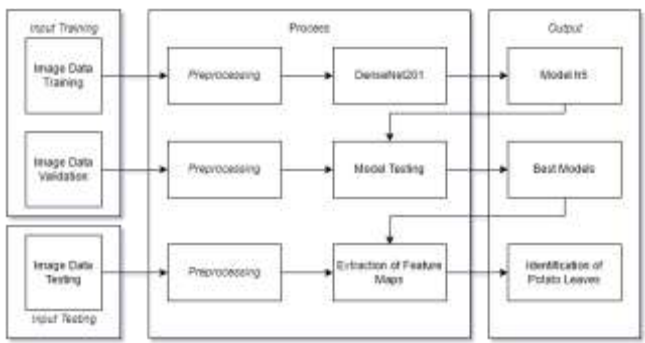

**Fig. 2** General Stages of the Classification Process

In the input training, the training stages are carried out in the training data with data validation. In the process stage, it performs preprocessing, such as resizing it to 224x224 pixels and normalizing the data after resizing it [16][17]. The training data image uses a process with the DenseNet model, which includes feature extraction and classification processes. The DenseNet model has common operations of batch normalization, ReLU activation, and convolution. The DenseNet model with 201 layers has processes of dense block 1, transition layer 1, dense block 2, transition layer 2, dense block 3, transition layer 3, dense block 4, and classification layer, which produces an output model with h5 format [18]. In the data validation process, model testing is carried out with training data which will produce output with the best model in terms of weight [19].

The testing data image enters into the preprocessing process by resizing it to 224x224 pixels and normalizing the data, then carrying out the feature maps extraction process by taking the best weight value from the training data so that it will produce identification output from potato leaf disease [20].

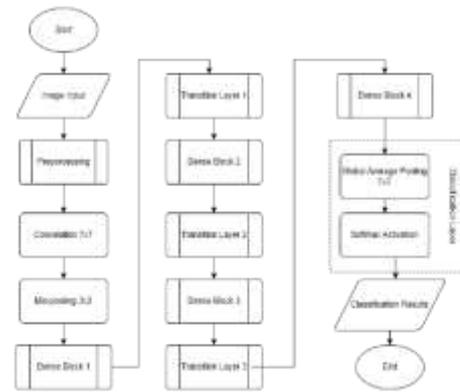

**Fig. 3** Densenet201 Algorithm Flowchart

Figure 3 illustrates the flowchart of the DenseNet201 algorithm in determining the identification of potato leaf disease. The stages of the DenseNet201 model begin with image input and preprocessing, then perform a 7x7 convolution operation with strides 2. Next, a 3x3 max pooling process with stides 2 will be carried out, which will later get a matrix value. This matrix value will be processed into a dense block 1 layer, transition layer 1, dense block 2, transition layer 2, dense block 3, transition layer 3, dense block 4. Matrix values will then be processed into the classification layer with 7x7 global average pooling operations and entered into softmax activation. Softmax activation is the last layer in the DenseNet201 model to determine the classification class [21].

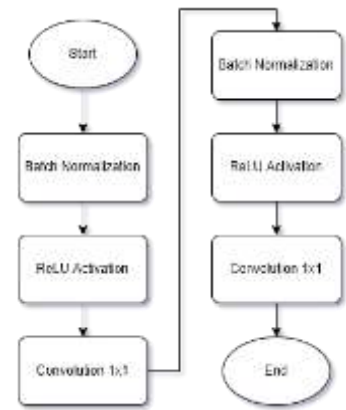

**Fig. 4** Denseblock

Figure 4 shows the stages of the process contained in the denseblock layer. These stages are batch normalization, ReLU activation, and convolution 1x1. This operation is also called a bottleneck. Then proceed with batch normalization, ReLU activation, convolution 3x3 where later the values of each matrix will be combined because they are interconnected. Inside the dense block layer, there are also multiplied operations so that it reaches 201 layers. In dense block 1 there is a convolution layer multiplied by 6. In dense block 2 there is a convolution layer multiplied by 12, the convolution layer in dense block 3 will be multiplied by 48, and the convolution layer in dense block 4 will be multiplied by 32 [22].

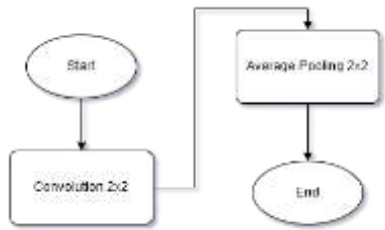

**Fig. 5** Transition Layer

Figure 5 shows the stages of the process contained in the transition layer. Some of these stages are 2x2 convolution operations and 1x1 average pooling with 2 strides. The transition layer will be placed after processing each dense block so that the number of transition layers becomes 3 [23].

*G. DenseNet201 Model Training and Testing*

At this stage, training and testing are carried out. This stage is carried out after the algorithm has been designed. Training and prediction testing of potato leaf disease classification were carried out using the DenseNet201 method in which there were several experiments:

*1)* The number of epochs is set to 100 to train the model further.

*2)* The number of batch sizes is set to 64 to make the computation process faster.

*3)* Comparing the optimizer Adam, SGD, and RMSprop

*4)* Compare dropouts with values 0.1, 0.2, 0.3, 0.4, 0.5, 0.6

*5)* Freeze the bottom layer on the DenseNet201 architecture, which is expected to prevent overfitting.

*6)* The proportion of data sharing used is 80% for training data and 20% for validation data.

*H. Analysis of Predictive Results*

The prediction result analysis stage checks the accuracy of the experimental results, which are designed into several test scenarios. The experimental results were then analyzed to conclude the analysis stage of the prediction results in this study using a confusion matrix to measure the performance of an algorithm [24].

*I. Conclusion*

At this stage, conclusions will be drawn to take the core of the research to become a complete and comprehensive understanding. In addition, the conclusion is also the answer to the formulation of the problem that has been made. If the research objectives and the final results are appropriate, this research can be used as a reference for further research. Suggestions that can be used to conduct further research related to the same problem as the related final project research can then also be written in the conclusions and suggestions sub-chapter [25].

### III. RESULTS AND DISCUSSION

In this study, researchers will classify three classes of potato leaf images, namely early blight, late blight, and healthy using Transfer Learning with the DenseNet201 architecture. This process begins with training on the dataset, which aims to form a model used in the testing process. The parameter that will be used to measure the model's performance is the accuracy value obtained from data testing.

*A. Preprocessing*

We will classify potato leaf disease using the DenseNet201 architecture in this study. Image of potato leaf disease obtained from the Kaggle website. This data consists of 3251 for training data and 416 for testing data. The potato leaf image used is a true color image that has three color channels, namely Red (R), Green (G), and Blue (B), and each channel has its pixel value. The original size of the potato leaf images is very different. Therefore, it is necessary to resize it so that each image has the same size. The image will be resized to 224x224 pixels, which is acceptable to the DenseNet201 architecture. After resizing, the data is multiplied for the training process using the FastTone Photo Resizer application so that the final number of images is 3900 for training data and 780 for testing data.

*B. DenseNet201 Model Design*

The DenseNet201 model is designed by calling the function from DenseNet, but it is set to false for the top layer because the researcher will add a separate layer. Then the researcher will adjust the size of the input image according to the size of the resized image, which is 224x224 pixels, and freeze the trainable layer so that the model does not experience overfitting, as shown in figure 6.

**Fig. 6** Calling the DenseNet201 Function

After defining the function, the researcher next makes an additional layer for the top layer. Researchers used 3 parameters for the top layer function: the base model, the number of neurons, and the activation function. For the first layer of the top layer, the researcher added the output from the base model, namely the DenseNet function, which was called earlier. The next layer is Global Average Pooling. The Global Average Pooling Layer aims to flatten before entering the fully connected layer. The size of the feature map matrix from the previous layer is then carried out by global average pooling by dividing the grid according to the size of the feature map. This layer does not add zero padding or stride.

The next layer is the dense layer. The dense layer itself is a layer in the architectural model that contains neurons. Technically the neurons in the neural network architecture work like the neurons in the human brain, which receive stimulants and process them to produce an output. In the top layer architecture, the dense layer is declared 5 times. The first dense layer contains the number of neurons multiplied by 3 and adds an activation function. The second dense layer contains the number of neurons multiplied by 1, and the activation function is added. The third dense layer contains the number of neurons multiplied by 2, and the activation function is added. The fourth dense layer contains the number of neurons multiplied by 1, and then the activation function is added. For the last dense layer, the number of neurons is divided by 2, and the activation function is added.

Each dense layer will be interspersed with dropout layers with a value of 0.1. Dropout is a layer used to prevent the model from experiencing overfitting. Dropouts are usually used for output on fully connected layers or neurons. Each iteration of the dropout training process will eliminate neurons randomly with a specified probability. The complete architecture of the top layer is depicted in figure 7.

**Fig. 7** Top Layer Architecture

After the top layer has been created, the next step is to call the top layer function to carry out the training process. The parameter of the top layer function contains the base model from DenseNet201, the number of neurons is set to 512, and the activation function used is ReLU so that the model does not experience backpropagation errors such as the sigmoid model. The top layer function call is declared, as shown in figure 8.

**Fig. 8** Calling the Top Layer Function

*C. DenseNet201 Model Training and Testing*

Model training and testing are done by comparing several trials to get the best results. The experiment was carried out by comparing the optimizer and dropout parameters. The optimizers used are Adam, RMSprop, and SGD, while the dropout layer parameters used are 0.1, 0.2, 0.3, 0.4, 0.5, and 0.6.

*1) Dropout Effect*

Dropout is a technique of removing some neurons randomly and will not be used during training. Dropout is a technique that can prevent overfitting and also speed up the training process. Dropout removes neurons in the hidden and visible layers of the network. If removing one neuron means removing that neuron from the existing randomly selected network. The probability value of each neuron is between 0 and 1. Dropout is a technique that is easy to apply in dealing with overfitting cases. The following is the result of the dropout parameter comparison.

**TABLE I.** Dropout Parameter Comparison Results

| Dropout | Training | Validation | Accuracy | Precision | Recall | F1-Score |
|---|---|---|---|---|---|---|
| 0.1 | 99.5 | 95.2 | 96 | 95.6 | 95.3 | 95.3 |
| 0.2 | 97.2 | 92.1 | 93 | 93 | 92.6 | 92.6 |
| 0.3 | 96.3 | 92.8 | 93 | 93 | 92.6 | 92.6 |
| 0.4 | 96.3 | 92.8 | 93 | 93 | 92.6 | 92.6 |
| 0.5 | 96 | 92.5 | 93 | 92.6 | 92.6 | 93 |
| 0.6 | 92.5 | 91 | 91 | 90.6 | 91 | 90.3 |

Based on table 1, it can be concluded that the optimal dropout parameter is at a value of 0.1 because it can produce a training accuracy of 99.5%, a validation accuracy of 95.2%, and an accuracy of the confusion matrix of 96%. As seen from table 4.1, when the value of the dropout parameter is greater, the accuracy results tend to decrease. It is caused by the increasing number of layers that are disabled, affecting the accuracy value obtained. In addition, the dropout also affects the pattern of the training graph, along with the results of the training graph, by comparing the dropout parameters.

The results of testing for testing data are also different. Here are the results of testing for data testing. The testing data used is 40 images and the images are taken randomly from datasets that the model has never recognized.

**TABLE II.** Model Predictions

| Dropout | Number of Correct Predictions |
|---|---|
| 0.1 | 37 |
| 0.2 | 35 |
| 0.3 | 37 |
| 0.4 | 36 |
| 0.5 | 36 |
| 0.6 | 36 |

Table 2 shows that the highest number of correct predictions is in the dropout with parameter 0.1, even though the loss graph shows overfitting. Still, in the classification process, the dropout value of 0.1 achieves good results. Figure 9 is a sample image prediction visualization from a dropout value of 0.1.

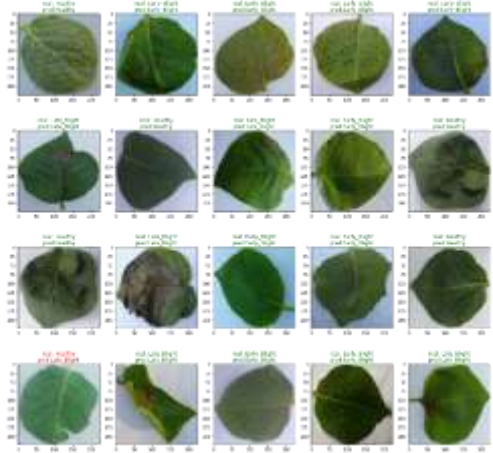

**Fig. 9** Sample Prediction Results Visualization

*2) Optimizer Effect*

Like other algorithms, neural network-based algorithms also have weaknesses, such as frequent delays in the convergence rate of data processing. Therefore, an optimization function is needed to prevent overfitting. Adam, SGD, and RMSprop optimization functions are the 3 best activation functions in a neural network when viewed from the perspective of low-loss training, testing, and validation in classification problems. Therefore the researcher will compare the 3 types of optimization functions in this study. The following table compares the results of the Adam, SGD, and RMSprop optimizers.

TABLE III. Optimizer Comparison Results

| Optimizer | Training | Validation | Accuracy | Precision | Recall | F1-Score |
|---|---|---|---|---|---|---|
| Adam | 99.5 | 95.2 | 96 | 95.6 | 95.3 | 95.3 |
| SGD | 98.5 | 93.4 | 93 | 93.3 | 93.3 | 94.3 |
| RMSprop | 98.6 | 94.4 | 95 | 94.6 | 94.6 | 94.3 |

Table 3 shows that the Adam parameter has good results regarding training accuracy of 99.5%, validation accuracy of 95.2%, and confusion matrix accuracy of 96%. Adam produces the highest accuracy because Adam is a combination of RMSprop and Stochastic Gradient Descent with momentum where Adam applies the adaptive learning rate method, namely calculating individual learning rate values for different parameters, and Adam uses the first and second-moment gradient estimates to adapt the learning rate for each weight of the neural network.

TABLE IV. Prediction Results

| Optimizer | Number of Correct Predictions |
|---|---|
| Adam | 3 |
| RMSprop | 3 |
| SGD | 6 |

Table 4 shows the optimization function adam and RMSprop predicting 37 correct images while SGD only predicts 34 correct images. Figure 10 is a sample image prediction visualization from the adam optimization function.

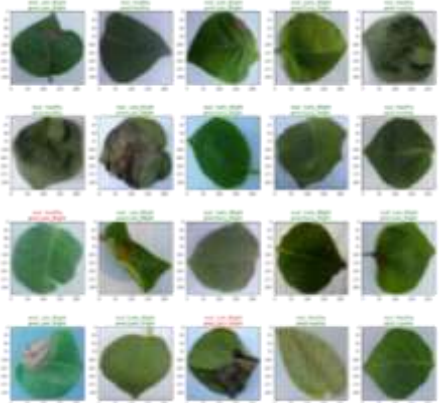

**Fig. 10** Sample Prediction Results Visualization

*D. Discussion*

Research discussing the classification of potato leaf disease was carried out by [9] using the VGG19 method to extract the relevant features from the dataset. The research resulted in an accuracy of 97.8%. The second study discussing the classification of potato leaf disease was carried out by [10] using the ResNet-152 and InceptionV3 methods. This research resulted in accuracy for ResNet-152 of 98.34% and InceptionV3 of 95.24% with a learning rate parameter of 0.0005.

Based on the results in this study with the same case study, the accuracy results were increased compared to previous studies of 99.5% for training, 95.2% for validation, and 96% for the confusion matrix using potato leaf image data totaling 3900 images with the DenseNet201 algorithm. This study's best parameter tuning results use a combination of a dropout value of 0.1 and the type of Adam's optimization function. The DenseNet201 model with optimizer = Adam and dropout = 0.1 achieved a high classification accuracy in potato leaf disease classification because the Adam optimizer is a method for stochastic optimization. It's based on gradient descent. It's computationally efficient and well-suited for problems with large data sets or many parameters. Adam optimizer can adjust the learning rate adaptively for each parameter. It allows the model to converge quickly to a good solution.

Dropout is a regularization technique that prevents overfitting by randomly disabling a certain percentage of neurons during training. In this case, dropout = 0.1 means that only 10% of the neurons will be disabled during training. It helps make the model more robust by reducing the number of neurons updated during training, thus reducing the chances of overfitting.

The combination of Adam optimizer and dropout = 0.1 helped the DenseNet201 model generalize well on the unseen data, which increased the model's accuracy. The DenseNet201 model was able to learn features that are generalizable to new data. The Adam optimizer helped converge quickly to a good solution, and the dropout helped to prevent overfitting. Based on these results, the proposed Transfer Learning method with DenseNet201 architecture can perform better classification based on a large number of datasets.

## CONCLUSION

Based on this study regarding the classification of potato leaf disease using the DenseNet201 architecture, it can be concluded that the scenario using the dropout value obtained the best accuracy level using a dropout of 0.1, namely 99.5% for training, 95.2% for validation, and 96% for the results of the confusion matrix. While the scenario using the optimizer type obtained the best accuracy using the Adam optimizer, namely 99.5% for training, 95.2% for validation, and 96% for the confusion matrix results.

The advantage of this research is that the selection of the deep learning method with the DenseNet201 architecture can be proven to increase the accuracy of the classification of potato leaf disease. The disadvantage of this research is that it only tests two scenarios, namely the number of dropouts and the optimizer, so it cannot know whether other parameters can improve classification accuracy. The researcher provides several suggestions for further research, which are expected to increase the number of classes in the classification of potato leaf disease and add parameters such as comparing several image input values, learning rate values, and the number of batch sizes so that it can produce a model with the best parameter tuning.